\documentclass[letterpaper, 10 pt, conference]{ieeeconf}  

\IEEEoverridecommandlockouts                              

\usepackage{cite}
\usepackage{amsmath,amssymb,amsfonts}
\usepackage{algorithmic}
\usepackage{graphicx}
\usepackage{textcomp}
\usepackage{xcolor}
\usepackage{booktabs}
\usepackage{pgfplots}
\usepackage{tikz}
\usepackage[implicit=false,colorlinks=true,hidelinks,urlcolor=black,bookmarks=false]{hyperref}
\def\BibTeX{{\rm B\kern-.05em{\sc i\kern-.025em b}\kern-.08em
		T\kern-.1667em\lower.7ex\hbox{E}\kern-.125emX}}

\title{\LARGE \bf
Improving the Intelligent Driver Model by Incorporating Vehicle Dynamics: Microscopic Calibration and Macroscopic Validation
}

\author{Dominik Salles$^{1}$, Steve Oswald$^{2}$ and Hans-Christian Reuss$^{2}$
\thanks{The research leading to these results is part of the project Artificial Intelligence And Mobility (AIAMO), which is funded by the German Federal Ministry for Digital and Transport (BMDV) with a total of 16.7 million euros.}
\thanks{$^{1}$Dominik Salles is with the Research Institute of Automotive Engineering and Vehicle Engines Stuttgart (FKFS), 70569 Stuttgart, Germany {\tt\small dominik.salles@fkfs.de}}%
\thanks{$^{2}$Steve Oswald and Hans-Christian Reuss are with the Institute of Automotive Engineering (IFS), University of Stuttgart, 70569 Stuttgart, Germany}%
}

\newcommand\copyrighttext{%
	\footnotesize \textcopyright \the\year{} IEEE. Personal use of this material is permitted. Permission from IEEE must be obtained for all other uses, including reprinting/republishing this material for advertising or promotional purposes, collecting new collected works for resale or redistribution to servers or lists, or reuse of any copyrighted component of this work in other works.}

\newcommand\copyrightnotice{%
	\begin{tikzpicture}[remember picture,overlay]
		\node[anchor=south,yshift=15pt] at (current page.south) {\fbox{\parbox{\dimexpr0.8\textwidth-\fboxsep-\fboxrule\relax}{\copyrighttext}}};
	\end{tikzpicture}%
}

\begin{document}

\maketitle
\thispagestyle{empty}
\pagestyle{empty}

\begin{abstract}
Microscopic traffic simulations are used to evaluate the impact of infrastructure modifications and evolving vehicle technologies, such as connected and automated driving. Simulated vehicles are controlled via car-following, lane-changing and junction models, which are designed to imitate human driving behavior. However, physics-based car-following models (CFMs) cannot fully replicate measured vehicle trajectories. Therefore, we present model extensions for the Intelligent Driver Model (IDM), of which some are already included in the Extended Intelligent Driver Model (EIDM), to improve calibration and validation results. They consist of equations based on vehicle dynamics and drive off procedures. In addition, parameter selection plays a decisive role. Thus, we introduce a framework to calibrate CFMs using drone data captured at a signalized intersection in Stuttgart, Germany. We compare the calibration error of the Krauss Model with the IDM and EIDM. In this setup, the EIDM achieves a $17.78\,\%$ lower mean error than the IDM, based on the distance difference between real world and simulated vehicles. Adding vehicle dynamics equations to the EIDM further improves the results by an additional $18.97\,\%$. The calibrated vehicle-driver combinations are then investigated by simulating the traffic in three different scenarios: at the original intersection, in a closed loop and in a stop-and-go wave. The data shows that the improved calibration process of individual vehicles, openly available at \href{https://www.github.com/stepeos/pycarmodel_calibration}{https://www.github.com/stepeos/pycarmodel\_calibration}, also provides more accurate macroscopic results.
\end{abstract}


\section{Introduction}
Improving traffic infrastructure is often limited by funding and spacial constraints. Simulations can be highly beneficial in the investigation of traffic measures, particularly in areas where modifications would imply interference in traffic. New intelligent systems can then be tested in any type of traffic situation and the performance examined beforehand. \copyrightnotice

In order to evaluate simulation results quantitatively, the local traffic conditions must first be understood and replicated. This consists of converting and building the street network, traffic light programs, elevation data and traffic demand, then importing the data into a traffic simulation software. When individual vehicle data are of particular interest, then a microscopic simulation such as Simulation of Urban Mobility (SUMO) \cite{Lopez.2018} is required. Dynamic objects (cars, trucks, buses, cyclists, pedestrians, etc.) interact via car-following \cite{Krauss.1997,Treiber.2000,Salles.2022}, lane-changing \cite{Erdmann.2015,Kesting.2007}, junction \cite{Krajzewicz.2013} and other models. These need to be calibrated to accurately measure the advantage of new intelligent systems with regard to energy consumption, emissions and traffic flow.

CFMs can be physics-based and data-driven, or a combination of both. In particular, models based on neural networks (NNs) have gained interest, as they outperform traditional CFMs such as the IDM in terms of single vehicle calibration \cite{Zhou.2017, Yang.2022, Naing.2023, Wang.2018, Zhu.2018}. Some studies combine the IDM with a NN to improve the results \cite{Mo.2021, Naing.2022}. Nevertheless, data-driven models still have drawbacks. The lack of data often leads to researchers only training one model for all vehicles \cite{Zhou.2017, Yang.2022, Naing.2023, Naing.2022}, while the IDM can be calibrated for each vehicle. In addition, the trained NN may perform poorly in unseen scenarios and cause collisions. To this date, no sophisticated traffic simulator for data-driven CFMs exists, which makes their macroscopic validation difficult.

Therefore, our main objective is to improve, calibrate and validate existing physics-based models to simulate heterogeneous driver behavior that represents the traffic for an entire city. Model improvements are based on vehicle dynamics, which the IDM lacks \cite{Fadhloun.2015}.

After a short review in section 2, the study focuses on three aspects of CFM calibration: the model itself, the optimization algorithm and the acquisition of trajectory data. More precisely, we introduce the calibration of the EIDM with the differential evolution (DE) algorithm \cite{Storn.1997}, using data from a drone camera to calibrate a multitude of vehicles at once. In the last section, the proposed calibration and model improvements are validated through three different traffic scenarios.

\section{Related Work}

Before calibration, a CFM needs to be selected or developed to accurately replicate the observed human driving behavior. Various physics-based CFM's have emerged from previous research. They can be categorized as \cite{Saifuzzaman.2014}: safety distance models (Gipps \cite{Gipps.1981}, Krauss \cite{Krauss.1997}), cellular automata (Nagel und Schreckenberg \cite{Nagel.1992}), optimal velocity models \cite{Bando.1995,Jiang.2001}, desired measures models (IDM \cite{Treiber.2000} and its extensions \cite{Treiber.2006,Treiber.2013b,Kesting.2010,Treiber.2017}), Gazis-Herman-Rothery models \cite{Gazis.1961}, models with perception thresholds (Wiedemann \cite{Wiedemann.1974}) as well as models based on various human aspects.
The primary model used in our work is the EIDM \cite{Salles.2022}, which was previously added to SUMO and combines the Improved IDM with equations for human driving behavior, creating a universal driver model. Additionally, a startup equation and a variable maximum acceleration parameter lead to more realistic acceleration patterns. To verify the benefits of the added equations, the EIDM is compared to the original IDM and to SUMO's default option, the Krauss model, in this work. The models and underlying mathematical expressions can be found in the provided literature and SUMO's software code.

The real world data for the calibration stem from aerial measurements in Stuttgart, Germany \cite{Salles.2022}. The trajectories used for the study were derived from vehicles coming to a full stop before driving over the signalized intersection. Each trajectory is approximately $200\,m$ long. Similar drone data from different locations has been published before, from highways \cite{Krajewski.2018}, roundabouts \cite{Krajewski.2020,Breuer.2020}, intersections \cite{Bock.2020, Zheng.2023}, highway ramps \cite{Moers.2022} and even large areas \cite{Barmpounakis.2020}. Comparable trajectories have also been extracted from stationary cameras \cite{U.S.DepartmentofTransportationFederalHighwayAdministration.2017,Klitzke.2022} and specially equipped vehicles \cite{Zhu.2018b,He.2023,Pourabdollah.2017,Punzo.2005,Makridis.2021}. However, drone data has the advantage to measure trajectories without affecting driver behavior and can simultaneously record multiple vehicles without occlusion. Although intra-driver variability cannot be captured due to the short trajectories, the data is perfectly suited to add inter-driver variability to the models. This means that the calibration approach can extract multiple parameter sets from the data, but they do not vary over time.


The algorithms used to minimize the error between the CFM and real-world data can be divided into local and global calibration approaches \cite{Treiber.2013}. Local approaches optimize the parameters for each data point and aggregate the results, e.g., using a particle filter \cite{Hoogendoorn.2006} or maximum likelihood estimation \cite{Hoogendoorn.2010,Treiber.2013}, while global approaches calibrate the model by iteratively improving the simulated trajectory. After each model run, the error between simulated and real world trajectory is used to refine the parameter sets. In this study, we use the global method in form of evolutionary algorithms. Previous work has focused on the genetic algorithm (GA) \cite{Jin.2014, Kesting.2008, Hamdar.2015, Zhu.2018b, He.2023, Pourabdollah.2017, DaVieiraRocha.2015}, but also includes the least squares method \cite{Treiber.2013}, downhill simplex \cite{Brockfeld.2004}, differential evolution \cite{Souza.2021} and dividing rectangles \cite{Li.2016} to solve the nonlinear optimization problem in CFM calibration.

\section{Model Calibration}

A microscopic traffic simulation can only perform well if its CFM parameters are selected appropriately. Therefore, a model calibration is necessary for each specific traffic scenario. This study aims to calibrate vehicle-driver combinations driving off from a traffic light using drone recordings.

The data set contains more than just drive-offs, but only a subset of those are suitable for calibration. First, all vehicles in the first row at the traffic light and all subsequent pairs of leader and follower are chosen. The former is termed free leader. Next, the candidates are filtered by the proposed selection criteria below, imposed on their trajectories. The goal is to keep only those vehicles that were recorded and tracked without error and that are suitable for the calibration process (e.g., omitting those with lane changes, as we calibrate car-following behavior).

\begin{enumerate}
	\item Extract vehicles from selected lanes only.
	\item Remove vehicles where either the leader or follower was incorrectly tracked. 
	\item Extract only vehicles that cross the intersection once.
	\item Extract only vehicles that come to a full stop.
	\item Remove all vehicles with lane changes.
	\item Remove leader and follower vehicles with negative distances (bumper to bumper).
\end{enumerate}

Although three lanes run from northeast to southwest through the intersection, only the two inner lanes are calibrated, resulting in 486 trajectories. The third lane is disregarded, because it includes trucks, right turning vehicles and occlusions by a tree.

\begin{figure}[htbp]
	\centerline{\includegraphics{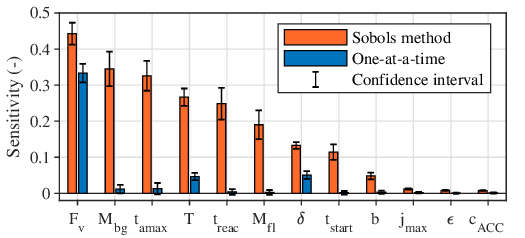}}
	\caption{Results of the sensitivity analysis with the EIDM parameters}
	\label{sensitivity}
\end{figure}

Before calibrating the models, we limit the number of CFM parameters to accelerate the optimization process. At first, stochastic parameters and properties are neglected, because they randomly change the outcome of every iteration and prevent the metaheuristic optimization from converging, especially if each parameter set is only calculated once. Hence, the parameters that enable stochastic model output are set to zero.
To investigate the importance of other parameters, a sensitivity analysis (SA) is performed. In the literature, two main approaches can be found: one-at-a-time SA \cite{Ciuffo.2014} and variance-based SA \cite{Kesting.2008}. We use both methods to analyze the sensitivity of the EIDM parameters. The maximum acceleration parameter $a_{\text{max}}$ is not included in the analysis, as it has by far the highest sensitivity value and a negative impact on the overall results, which are shown in Fig.\,\ref{sensitivity}. In our case, Sobol's method \cite{Sobol.2001,Saltelli.2010}, a variance-based SA, is superior to one-at-a-time SA, because it varies the values of multiple parameters at the same time to derive the total order sensitivity of each parameter. One-at-a-time SA may over- or underestimate the importance of parameters due to their mutual influence on the results, e.g., the IDM parameters $\delta$ and $a_{\text{max}}$ affect each other's sensitivity with regard to the model output.

As a result of the SA, the parameters $a_{\text{max}}$, time headway $T$ and speed factor $F_{\text{v}}$ are calibrated for all models, the desired deceleration $b$ and action step length $t_{\text{AP}}$ for Krauss and IDM and the acceleration exponent $\delta$ for IDM and EIDM. The EIDM calibration includes four additional parameters: the reaction time $t_{\text{reac}}$, start-up delay time $t_{\text{start}}$, start-up begin factor $M_{\text{bg}}$ and the time until maximal acceleration $t_{\text{amax}}$.
This shows that omitting parameters is scenario-specific, e.g., given the drive-off data set from \cite{Salles.2022}, the impact of the deceleration parameter on the result of the EIDM is marginal.
The calibration procedure is divided into four steps:

\begin{enumerate}
	\item Preprocess the data (load drone data, calculate distance to lanes, etc.).
	\item Perform the selection process using the aforementioned criteria.
	\item Start the traffic simulation.
	\item For every calibration candidate:
	\begin{itemize}
		\item Do an estimation of the parameters.
		\item Perform optimization of the CFM parameters.
	\end{itemize}
\end{enumerate}

According to \cite{Punzo.2021}, the optimization process can generally be described as follows:
\begin{equation}
	MoP^{sim} = F(\beta)
\end{equation}
\begin{equation}
	min\ GoF(MoP^{gt}, MoP^{sim})
\end{equation}
\begin{equation}
	subject\ to\  \text{LB}_{\beta}\leq \beta \leq \text{UB}_{\beta}
\end{equation}

In this context, $MoP^{sim}$ refers to the measure of performance of the simulation output $F$ with parameter set $\beta$. We select the single $MoP$ distance between leader and follower (bumper to bumper) as optimal, based on conclusions of previous studies \cite{Punzo.2021,Treiber.2013}. $GoF$ stands for goodness-of-fit and quantifies the error between the simulated output and the ground truth data, as evaluated by $MoP$. It is calculated as root-mean-square error (RMSE), as proposed in \cite{Punzo.2021}.
In pursuit of the best possible solution, the following steps are performed to calculate the $GoF$ for the optimization:

\begin{enumerate}
	\item Initialize a population of size $N$ with random parameters between upper and lower boundaries.
	\item Set the first population instance with the estimated parameters.
	\item Build a network with $N$-amount of lanes and initialize the traffic simulation.
	\item For every optimization: Insert leader and follower vehicles into the simulation. Then, set the follower's parameter sets from the optimization algorithms' solutions. Next, replicate leader-vehicle behavior from ground truth data using their driven distance. Lastly, calculate the $GoF$ from the simulation output.
\end{enumerate}

By simulating multiple solutions in one model run, the calibration process is parallelized and sped up significantly. In addition, every implemented model in SUMO can be calibrated, as the followers are moved by SUMO and the leader vehicles positioned using the ground truth. To avoid lane changes and interference between the population in the simulation, each pair drives in a single-lane.

To compare optimization algorithms, we calibrate the same data with two different metaheuristic optimization algorithms. Next to the DE algorithm with a population size of 200 for 50 iterations, we use the GA for 33 candidates with a population size of 500 for 1000 iterations. This increases the certainty of the GA for reaching the global optimum. After 50 iterations, the DE has a lower error value in 28 out of the 33 cases, while the GA outperforms the 50-iteration-DE after an average of 174 iterations. The GA outperforms the $GoF$ results of the DE by an average of $8.6\,\%$. The DE algorithm is observed to be far more robust to converge, despite changing the hyperparameters, while the GA takes longer. Based on these findings, the DE is selected to calibrate the different CFMs.

When choosing an optimization algorithm, time is the dominating factor. We calibrate the data on an Intel(R) Core(TM) i9-10900X with a time step of $0.04\,s$. One iteration takes an average of $35\,s$ with a population size of 200, $100\,s$ with a population size of 500 and about $150\,s$ with a size of 1000, independent of the optimization algorithm. The calibration time also increases with the candidate's captured trajectory time, which is only around $25\,s$ in the data set used here.


\section{Model Validation}

In this section, the calibration results of the Krauss model, IDM and EIDM are analyzed and compared to simulations with SUMO's default car-following parameters of the same models. Two additional EIDM configurations complete the analysis. They include the new speed dependent acceleration parameter, which turns $a_{\text{max}}$ into a vector. Each value is calibrated separately and applied to a specific velocity range, which is indicated in the model name (e.g., EIDM\_5\_12 means that $a_{\text{max}}$ consists of two fixed values for velocities between 0 and $5\,m/s$ and above $12\,m/s$, with a linear interpolation in between).

We run three different simulations per configuration, using a modified SUMO version 1.16. The goal is to relate the microscopic (e.g., acceleration comparison of a single vehicle) to the macroscopic results (e.g., traffic flow). For the correct representation of a traffic network, both entities need to deliver appropriate results.

\begin{figure}[bp]
	\centerline{\includegraphics{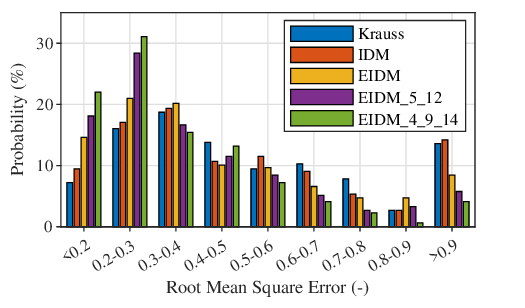}}
	\caption{RMSE between the simulated trajectory with the calibrated model and the corresponding real world trajectory}
	\label{calib_results}
\end{figure}

First, we compare the RMSE values of the calibration process. Fig.\,\ref{calib_results} summarizes their distributions across 486 calibrated parameter sets. The EIDM produces the best results, which is mainly due to the startup equation and delay. Further improvements over EIDM's static $a_{\text{max}}$ value are achieved with the velocity dependent acceleration parameter.

\begin{figure*}[htbp]
	\includegraphics{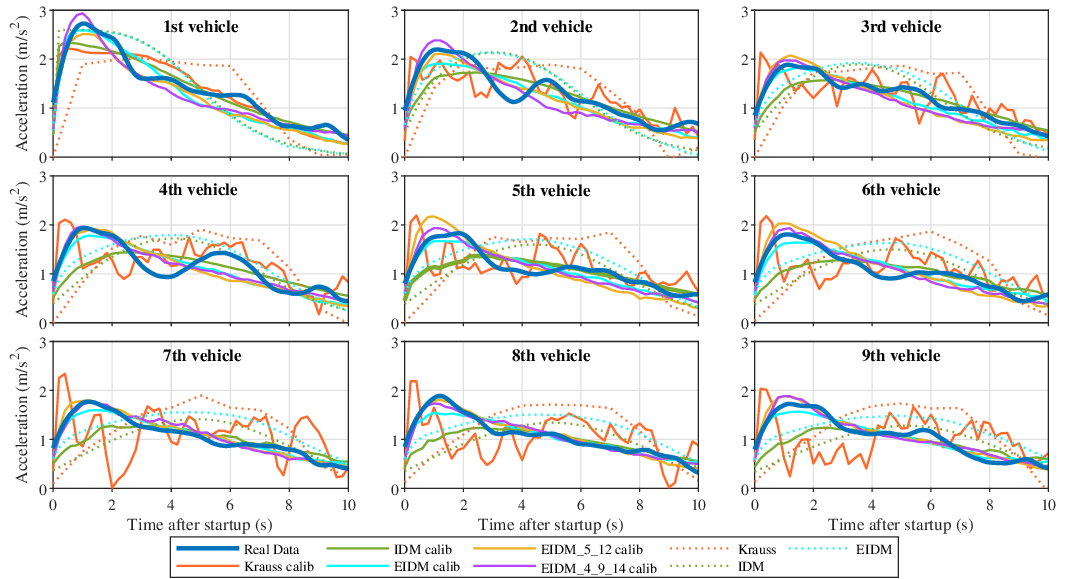}
	\caption{Average acceleration curves of the first nine vehicles queued at the signalized intersection}
	\label{accel_in_row}
\end{figure*}

\subsection{Original Intersection Simulation}

These findings are evident in Fig.\,\ref{accel_in_row}. The graphs show the acceleration curves of the first nine vehicles after startup following a free leader. The average results are displayed in separate graphs to highlight the performance of the individual models. The simulations were carried out with SUMO and a map of the exact intersection where the drone data was captured \cite{Salles.2022}. All models simulate the behavior of the first vehicle in the queue fairly well. After that, the results start to differ. In homogeneous traffic, the vehicles with default parameters start to accelerate slowly and reach their maximum after approx. $5\,s$. By calibrating the model parameters, the maximum acceleration is shifted to an earlier point in time, which coincides with observations. The Krauss model is an exception, since it possesses two unique features, which make it difficult to create heterogeneous traffic. Firstly, different deceleration parameters $\mathit{b}$ alter the behavior of following vehicles and, secondly, the desired headway $\mathit{T}$ must be higher than the simulation time step in order to avoid accidents. Independent of $\mathit{T}$, high $t_{\text{AP}}$ values (SUMO parameter) always result in improper traffic flow. The first issue is solved in SUMO by letting each driver know the desired maximum deceleration of the driver in front. Therefore, the driver can keep a greater distance and decelerate earlier. In reality, the driver does not have this information of the preceding vehicle, although this could be anticipated.

\begin{table}[!bp]
	\caption{Simulation results using the calibrated models and those with default parameters}
	\begin{center}
		\begin{tabular}{l c c c c}
			\toprule
			\textbf{Model}&\parbox{1.1cm}{\centering\textbf{Number of Collisions}}&\parbox{0.9cm}{\centering\textbf{Number of} \\ \textbf{Emerg. stops}}&\parbox{0.9cm}{\centering\textbf{Average} \\ \textbf{RMSE}}&\parbox{1.4cm}{\centering\textbf{Average number of} \\ \textbf{vehicles }\\ \textbf{per lane and cycle}} \\
			\midrule
			Real data & 0 & 0 & 0.0000 & 17.50 \\
			Krauss calib & 13829 & 6 & 0.5952 & 17.43 \\
			IDM calib & 36 & 4 & 0.5858 & 18.97 \\
			EIDM calib & 0 & 0 & 0.4817 & 16.27 \\
			EIDM\_5\_12 calib & 0 & 0 & 0.4136 & 16.581 \\
			EIDM\_4\_9\_14 calib & 0 & 0 & 0.3706 & 16.27 \\
			Krauss & 0 & 0 & - & 20.37 \\
			Krauss step 1s & 0 & 0 & - & 16.93 \\
			Krauss Action 1s & 7587 & 48 & - & 14.77 \\
			IDM & 0 & 0 & - & 16.48 \\
			EIDM & 0 & 0 & - & 17.18 \\
			\bottomrule
		\end{tabular}
		\label{tab_results}
	\end{center}
\end{table}

Table\,\ref{tab_results} shows the average calibrations results over all 486 calibrated vehicle-driver combinations. Additionally listed are the amount of collisions, emergency stops and average vehicles per traffic light cycle during the $2600\,s$ (43 green phases) SUMO simulation at the intersection in Stuttgart. The number of collisions simulated by the Krauss model and the lower RMSE values of the EIDM confirm the findings mentioned above.

Based on these results, the EIDM is the most suitable model to recreate the measured acceleration patterns. However, to accurately reproduce traffic, macroscopic values must also coincide with measured ones, such as traffic flow, characteristic wave velocity and travel times.

\begin{figure}[htbp]
	\includegraphics{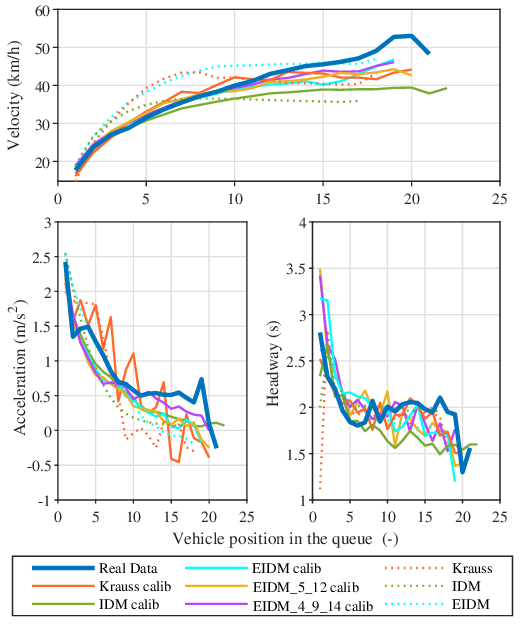}
	\caption{Real data and model results for time headway, velocity and acceleration of the queued vehicles after crossing the stop line}
	\label{velo_accel_time}
\end{figure}

First, we compare the simulated average acceleration, velocity and headway with the real data after crossing the stop line. The data was captured with simulations at the replicated intersection and is shown in Fig.\,\ref{velo_accel_time}. The calibrated EIDM shows the highest resemblance to the real data with one main discrepancy. After the 10th vehicle, the simulated velocity, acceleration and headway of the following vehicles start to deviate from the real data. This results from the measurement process and the field of view (FOV) of the drone camera. The first vehicles reach their final speed after driving out of view and vehicles that did not stop are excluded from the calibration process. Therefore, trajectories near the speed limit are underrepresented. Lower velocities result in fewer vehicles crossing the intersection per traffic light cycle than observed, as shown for each model in Table\,\ref{tab_results}, unless bumper to bumper distances are low.

\subsection{Closed Loop Simulation}

\begin{figure}[htbp]
	\includegraphics{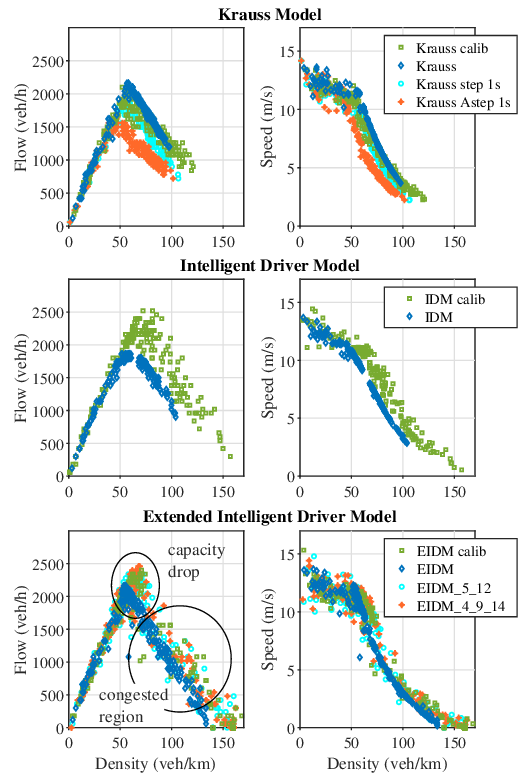}
	\caption{Flow-density and speed-density diagrams of the circular route simulations}
	\label{fundamental_diagram}
\end{figure}

In addition to the intersection model, we create a closed loop simulation. The loop is $3.4\,km$ long with a speed limit of $50\,km/h$ and consists of three measurement zones of $50\,m$ each. Each simulation lasts $4200\,s$, during which a vehicle is inserted every $5\,s$ if there is enough space. To create stop-and-go waves, we close a short road segment every $250\,s$ for $25\,s$, which leads to the results in Fig.\,\ref{fundamental_diagram}. It shows that the traffic flow generated by the calibrated, heterogeneous EIDM exhibits the capacity drop observed in real traffic \cite{Treiber.2000}, before slowly decreasing. In case the magnitude of the simulated capacity drop needs to be increased, the headway parameter $T$ can be converted into a velocity dependent vector \cite{Treiber.2013b}. In this setup, only the EIDM produces stop-and-go waves. Nevertheless, both the EIDM and the calibrated IDM generate diverse data points in the region of congested traffic, which is a measurable phenomenon in road traffic.

\subsection{Stop-and-go Simulation}

\begin{figure*}[htbp]
	\includegraphics{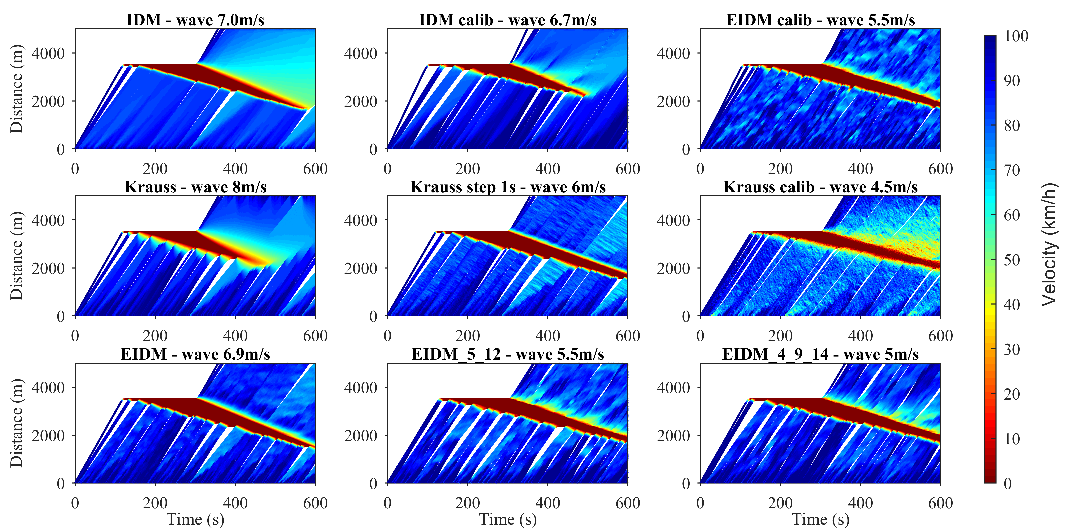}
	\caption{Distance-time graphs of stop-and-go waves using different models}
	\label{traffic_wave}
\end{figure*}

The last simulation shows model differences with regard to stop-and-go waves. It consists of a $5\,km$ long, straight road with a temporary road closure after $3.5\,km$. When the road is reopened, a moving traffic wave (congestion front) can be measured. Distance-time graphs of each simulation are shown in Fig.\,\ref{traffic_wave}. The traffic wave is different for each model and parameter set, while most of the presented combinations are overestimating the characteristic wave velocity of $15-20\,km/h$ ($4.17-5.56\,m/s$) \cite{Kerner.1996,Treiber.2013b}. In contrast to other studies, we are using data from inner city traffic and not from congested highways, which could be the reason for the performance gap.

\section{Conclusion}

This study presents a framework to calibrate various CFMs using SUMO and drone data. Next to the genetic algorithm, we introduce an optimization approach with the differential evolution algorithm, which outperforms the genetic algorithm. We compare different CFMs and show that the EIDM produces the lowest calibration errors, resulting in more realistic acceleration patterns and, therefore, velocities and distances. The model also replicates macroscopic traffic phenomena, such as characteristic wave velocities, a capacity drop and diverse data points in the congested traffic region. Whether the simulated values correspond to reality remains to be verified by further traffic measurements.

Our goal for the future is to calibrate the models using more data sets and longer trajectories, such as those published by \cite{Barmpounakis.2020}, to appropriately simulate velocities close to the speed limit. Macroscopic results may further improve with multi-objective optimization, which was previously added to the calibration framework. We also aim to extend the EIDM with a variable headway parameter $T$ specific to the traffic situation and velocity as well as a maximum acceleration calculation depending on vehicle resistance.

%
%
\bibliographystyle{IEEEtran}

\bibliography{IEEEabrv,Calibrating_Car-Following_IEEE_DSA}

\begin{thebibliography}{10}
\providecommand{\url}[1]{#1}
\csname url@samestyle\endcsname
\providecommand{\newblock}{\relax}
\providecommand{\bibinfo}[2]{#2}
\providecommand{\BIBentrySTDinterwordspacing}{\spaceskip=0pt\relax}
\providecommand{\BIBentryALTinterwordstretchfactor}{4}
\providecommand{\BIBentryALTinterwordspacing}{\spaceskip=\fontdimen2\font plus
\BIBentryALTinterwordstretchfactor\fontdimen3\font minus
  \fontdimen4\font\relax}
\providecommand{\BIBforeignlanguage}[2]{{%
\expandafter\ifx\csname l@#1\endcsname\relax
\typeout{** WARNING: IEEEtran.bst: No hyphenation pattern has been}%
\typeout{** loaded for the language `#1'. Using the pattern for}%
\typeout{** the default language instead.}%
\else
\language=\csname l@#1\endcsname
\fi
#2}}
\providecommand{\BIBdecl}{\relax}
\BIBdecl

\bibitem{Lopez.2018}
P.~A. Lopez, E.~Wiessner, M.~Behrisch, L.~Bieker-Walz, J.~Erdmann, Y.-P.
  Flotterod, R.~Hilbrich, L.~Lucken, J.~Rummel, and P.~Wagner, ``Microscopic
  traffic simulation using sumo.''\hskip 1em plus 0.5em minus 0.4em\relax IEEE,
  2018, pp. 2575--2582.

\bibitem{Krauss.1997}
S.~Krauss, P.~Wagner, and C.~Gawron, ``Metastable states in a microscopic model
  of traffic flow,'' \emph{Physical review. E, Statistical physics, plasmas,
  fluids, and related interdisciplinary topics}, vol.~55, no.~5, pp.
  5597--5602, 1997.

\bibitem{Treiber.2000}
Treiber, Hennecke, and Helbing, ``Congested traffic states in empirical
  observations and microscopic simulations,'' \emph{Physical review. E,
  Statistical physics, plasmas, fluids, and related interdisciplinary topics},
  vol.~62, no. 2 Pt A, pp. 1805--1824, 2000.

\bibitem{Salles.2022}
D.~Salles, S.~Kaufmann, and H.-C. Reuss, ``Extending the intelligent driver
  model in sumo and verifying the drive off trajectories with aerial
  measurements,'' \emph{SUMO Conference Proceedings}, vol.~1, pp. 1--25, 2022.

\bibitem{Erdmann.2015}
J.~Erdmann, ``Sumo's lane-changing model,'' in \emph{Modeling Mobility with
  Open Data}, ser. Lecture Notes in Mobility, M.~Behrisch and M.~Weber,
  Eds.\hskip 1em plus 0.5em minus 0.4em\relax Cham: {Springer International
  Publishing}, 2015, pp. 105--123.

\bibitem{Kesting.2007}
A.~Kesting, M.~Treiber, and D.~Helbing, ``General lane-changing model mobil for
  car-following models,'' \emph{Transportation Research Record: Journal of the
  Transportation Research Board}, vol. 1999, no.~1, pp. 86--94, 2007.

\bibitem{Krajzewicz.2013}
\BIBentryALTinterwordspacing
D.~Krajzewicz and J.~Erdmann, ``Road intersection model in sumo,'' in \emph{1st
  SUMO User Conference - SUMO 2013}, ser. Reports of the DLR-Institute of
  Transportation Systems, vol.~21.\hskip 1em plus 0.5em minus 0.4em\relax DLR,
  2013, pp. 212--220. [Online]. Available: \url{https://elib.dlr.de/84363/}
\BIBentrySTDinterwordspacing

\bibitem{Zhou.2017}
M.~Zhou, X.~Qu, and X.~Li, ``A recurrent neural network based microscopic car
  following model to predict traffic oscillation,'' \emph{Transportation
  Research Part C: Emerging Technologies}, vol.~84, pp. 245--264, 2017.

\bibitem{Yang.2022}
L.~Yang, S.~Fang, G.~Wu, H.~Sheng, Z.~Xu, M.~Zhang, and X.~Zhao, ``Physical
  model versus artificial neural network (ann) model: A comparative study on
  modeling car-following behavior at signalized intersections,'' \emph{Journal
  of Advanced Transportation}, vol. 2022, pp. 1--18, 2022.

\bibitem{Naing.2023}
H.~Naing, W.~Cai, J.~Yu, T.~Wu, and L.~Yu, ``Physics-guided graph convolutional
  network for modeling car-following behaviors under distributional shift,'' in
  \emph{2023 IEEE 26th International Conference on Intelligent Transportation
  Systems (ITSC)}.\hskip 1em plus 0.5em minus 0.4em\relax IEEE, 2023, pp.
  2574--2581.

\bibitem{Wang.2018}
X.~Wang, R.~Jiang, L.~Li, Y.~Lin, X.~Zheng, and F.-Y. Wang, ``Capturing
  car-following behaviors by deep learning,'' \emph{IEEE Transactions on
  Intelligent Transportation Systems}, vol.~19, no.~3, pp. 910--920, 2018.

\bibitem{Zhu.2018}
M.~Zhu, X.~Wang, and Y.~Wang, ``Human-like autonomous car-following model with
  deep reinforcement learning,'' \emph{Transportation Research Part C: Emerging
  Technologies}, vol.~97, pp. 348--368, 2018.

\bibitem{Mo.2021}
Z.~Mo, R.~Shi, and X.~Di, ``A physics-informed deep learning paradigm for
  car-following models,'' \emph{Transportation Research Part C: Emerging
  Technologies}, vol. 130, p. 103240, 2021.

\bibitem{Naing.2022}
H.~Naing, W.~Cai, T.~Wu, and L.~Yu, ``Dynamic car-following model calibration
  with deep reinforcement learning,'' in \emph{2022 IEEE 25th International
  Conference on Intelligent Transportation Systems (ITSC)}.\hskip 1em plus
  0.5em minus 0.4em\relax IEEE, 2022, pp. 959--966.

\bibitem{Fadhloun.2015}
K.~Fadhloun, H.~Rakha, A.~Loulizi, and A.~Abdelkefi, ``Vehicle dynamics model
  for estimating typical vehicle accelerations,'' \emph{Transportation Research
  Record: Journal of the Transportation Research Board}, vol. 2491, no.~1, pp.
  61--71, 2015.

\bibitem{Storn.1997}
R.~Storn and K.~Price, ``Differential evolution -- a simple and efficient
  heuristic for global optimization over continuous spaces,'' \emph{Journal of
  Global Optimization}, vol.~11, no.~4, pp. 341--359, 1997.

\bibitem{Saifuzzaman.2014}
M.~Saifuzzaman and Z.~Zheng, ``Incorporating human-factors in car-following
  models: A review of recent developments and research needs,''
  \emph{Transportation Research Part C: Emerging Technologies}, vol.~48, pp.
  379--403, 2014.

\bibitem{Gipps.1981}
P.~G. Gipps, ``A behavioural car-following model for computer simulation,''
  \emph{Transportation Research Part B: Methodological}, vol.~15, no.~2, pp.
  105--111, 1981.

\bibitem{Nagel.1992}
K.~Nagel and M.~Schreckenberg, ``A cellular automaton model for freeway
  traffic,'' \emph{Journal de Physique I}, vol.~2, no.~12, pp. 2221--2229,
  1992.

\bibitem{Bando.1995}
Bando, Hasebe, Nakayama, Shibata, and Sugiyama, ``Dynamical model of traffic
  congestion and numerical simulation,'' \emph{Physical review. E, Statistical
  physics, plasmas, fluids, and related interdisciplinary topics}, vol.~51,
  no.~2, pp. 1035--1042, 1995.

\bibitem{Jiang.2001}
R.~Jiang, Q.~Wu, and Z.~Zhu, ``Full velocity difference model for a
  car-following theory,'' \emph{Physical review. E, Statistical, nonlinear, and
  soft matter physics}, vol.~64, no. 1 Pt 2, p. 017101, 2001.

\bibitem{Treiber.2006}
M.~Treiber, A.~Kesting, and D.~Helbing, ``Delays, inaccuracies and anticipation
  in microscopic traffic models,'' \emph{Physica A: Statistical Mechanics and
  its Applications}, vol. 360, no.~1, pp. 71--88, 2006.

\bibitem{Treiber.2013b}
M.~Treiber and A.~Kesting, \emph{Traffic flow dynamics: Data, models and
  simulation}.\hskip 1em plus 0.5em minus 0.4em\relax Heidelberg and New York:
  Springer, 2013.

\bibitem{Kesting.2010}
A.~Kesting, M.~Treiber, and D.~Helbing, ``Enhanced intelligent driver model to
  access the impact of driving strategies on traffic capacity,''
  \emph{Philosophical transactions. Series A, Mathematical, physical, and
  engineering sciences}, vol. 368, no. 1928, pp. 4585--4605, 2010.

\bibitem{Treiber.2017}
M.~Treiber and A.~Kesting, ``The intelligent driver model with stochasticity
  -new insights into traffic flow oscillations,'' \emph{Transportation Research
  Procedia}, vol.~23, pp. 174--187, 2017.

\bibitem{Gazis.1961}
D.~C. Gazis, R.~Herman, and R.~W. Rothery, ``Nonlinear follow-the-leader models
  of traffic flow,'' \emph{Operations Research}, vol.~9, no.~4, pp. 545--567,
  1961.

\bibitem{Wiedemann.1974}
R.~Wiedemann, ``{Simulation des Straßenverkehrsflusses},'' Hochschulschrift,
  Karlsruhe, 1974.

\bibitem{Krajewski.2018}
R.~Krajewski, J.~Bock, L.~Kloeker, and L.~Eckstein, ``The highd dataset: A
  drone dataset of naturalistic vehicle trajectories on german highways for
  validation of highly automated driving systems.''\hskip 1em plus 0.5em minus
  0.4em\relax IEEE, 2018, pp. 2118--2125.

\bibitem{Krajewski.2020}
R.~Krajewski, T.~Moers, J.~Bock, L.~Vater, and L.~Eckstein, ``The round
  dataset: A drone dataset of road user trajectories at roundabouts in
  germany,'' in \emph{2020 IEEE 23rd International Conference on Intelligent
  Transportation Systems (ITSC)}.\hskip 1em plus 0.5em minus 0.4em\relax IEEE,
  2020, pp. 1--6.

\bibitem{Breuer.2020}
A.~Breuer, J.-A. Termohlen, S.~Homoceanu, and T.~Fingscheidt, ``opendd: A
  large-scale roundabout drone dataset,'' in \emph{2020 IEEE 23rd International
  Conference on Intelligent Transportation Systems (ITSC)}.\hskip 1em plus
  0.5em minus 0.4em\relax IEEE, 2020, pp. 1--6.

\bibitem{Bock.2020}
J.~Bock, R.~Krajewski, T.~Moers, S.~Runde, L.~Vater, and L.~Eckstein, ``The ind
  dataset: A drone dataset of naturalistic road user trajectories at german
  intersections,'' in \emph{2020 IEEE Intelligent Vehicles Symposium
  (IV)}.\hskip 1em plus 0.5em minus 0.4em\relax IEEE, 2020, pp. 1929--1934.

\bibitem{Zheng.2023}
O.~Zheng, M.~Abdel-Aty, L.~Yue, A.~Abdelraouf, Z.~Wang, and N.~Mahmoud,
  ``Citysim: A drone-based vehicle trajectory dataset for safety-oriented
  research and digital twins,'' \emph{Transportation Research Record: Journal
  of the Transportation Research Board}, 2023.

\bibitem{Moers.2022}
T.~Moers, L.~Vater, R.~Krajewski, J.~Bock, A.~Zlocki, and L.~Eckstein, ``The
  exid dataset: A real-world trajectory dataset of highly interactive highway
  scenarios in germany,'' in \emph{2022 IEEE Intelligent Vehicles Symposium
  (IV)}.\hskip 1em plus 0.5em minus 0.4em\relax IEEE, 2022, pp. 958--964.

\bibitem{Barmpounakis.2020}
E.~Barmpounakis and N.~Geroliminis, ``On the new era of urban traffic
  monitoring with massive drone data: The pneuma large-scale field
  experiment,'' \emph{Transportation Research Part C: Emerging Technologies},
  vol. 111, pp. 50--71, 2020.

\bibitem{U.S.DepartmentofTransportationFederalHighwayAdministration.2017}
\BIBentryALTinterwordspacing
{U.S. Department of Transportation Federal Highway Administration}, ``Next
  generation simulation (ngsim) vehicle trajectories and supporting data,''
  [Dataset]. Provided by ITS DataHub through Data.transportation.gov. [Online].
  Available: \url{http://doi.org/10.21949/1504477}
\BIBentrySTDinterwordspacing

\bibitem{Klitzke.2022}
L.~Klitzke, K.~Gimm, C.~Koch, and F.~Koster, ``Extraction and analysis of
  highway on-ramp merging scenarios from naturalistic trajectory data,'' in
  \emph{2022 IEEE 25th International Conference on Intelligent Transportation
  Systems (ITSC)}.\hskip 1em plus 0.5em minus 0.4em\relax IEEE, 2022, pp.
  654--660.

\bibitem{Zhu.2018b}
M.~Zhu, X.~Wang, A.~Tarko, and S.~Fang, ``Modeling car-following behavior on
  urban expressways in shanghai: A naturalistic driving study,''
  \emph{Transportation Research Part C: Emerging Technologies}, vol.~93, pp.
  425--445, 2018.

\bibitem{He.2023}
L.~He and X.~Wang, ``Calibrating car-following models on urban streets using
  naturalistic driving data,'' \emph{Journal of Transportation Engineering,
  Part A: Systems}, vol. 149, no.~4, 2023.

\bibitem{Pourabdollah.2017}
M.~Pourabdollah, E.~Bjarkvik, F.~Furer, B.~Lindenberg, and K.~Burgdorf,
  ``Calibration and evaluation of car following models using real-world driving
  data,'' in \emph{2017 IEEE 20th International Conference on Intelligent
  Transportation Systems (ITSC)}.\hskip 1em plus 0.5em minus 0.4em\relax IEEE,
  2017, pp. 1--6.

\bibitem{Punzo.2005}
V.~Punzo and F.~Simonelli, ``Analysis and comparison of microscopic traffic
  flow models with real traffic microscopic data,'' \emph{Transportation
  Research Record: Journal of the Transportation Research Board}, vol. 1934,
  pp. 53--63, 2005.

\bibitem{Makridis.2021}
M.~Makridis, K.~Mattas, A.~Anesiadou, and B.~Ciuffo, ``Openacc. an open
  database of car-following experiments to study the properties of commercial
  acc systems,'' \emph{Transportation Research Part C: Emerging Technologies},
  vol. 125, p. 103047, 2021.

\bibitem{Treiber.2013}
M.~Treiber and A.~Kesting, ``Microscopic calibration and validation of
  car-following models -- a systematic approach,'' \emph{Procedia - Social and
  Behavioral Sciences}, vol.~80, pp. 922--939, 2013.

\bibitem{Hoogendoorn.2006}
S.~Hoogendoorn, S.~Ossen, M.~Schreuder, and B.~Gorte, ``Unscented particle
  filter for delayed car-following models estimation,'' in \emph{2006 IEEE
  Intelligent Transportation Systems Conference}.\hskip 1em plus 0.5em minus
  0.4em\relax IEEE, 2006, pp. 1598--1603.

\bibitem{Hoogendoorn.2010}
S.~Hoogendoorn and R.~Hoogendoorn, ``Calibration of microscopic traffic-flow
  models using multiple data sources,'' \emph{Philosophical transactions.
  Series A, Mathematical, physical, and engineering sciences}, vol. 368, no.
  1928, pp. 4497--4517, 2010.

\bibitem{Jin.2014}
P.~J. Jin, {Da Yang}, and B.~Ran, ``Reducing the error accumulation in
  car-following models calibrated with vehicle trajectory data,'' \emph{IEEE
  Transactions on Intelligent Transportation Systems}, vol.~15, no.~1, pp.
  148--157, 2014.

\bibitem{Kesting.2008}
A.~Kesting and M.~Treiber, ``Calibrating car-following models by using
  trajectory data,'' \emph{Transportation Research Record: Journal of the
  Transportation Research Board}, vol. 2088, no.~1, pp. 148--156, 2008.

\bibitem{Hamdar.2015}
S.~H. Hamdar, H.~S. Mahmassani, and M.~Treiber, ``From behavioral psychology to
  acceleration modeling: Calibration, validation, and exploration of drivers'
  cognitive and safety parameters in a risk-taking environment,''
  \emph{Transportation Research Part B: Methodological}, vol.~78, pp. 32--53,
  2015.

\bibitem{DaVieiraRocha.2015}
T.~{Da Vieira Rocha}, L.~Leclercq, M.~Montanino, C.~Parzani, V.~Punzo,
  B.~Ciuffo, and D.~Villegas, ``Does traffic-related calibration of
  car-following models provide accurate estimations of vehicle emissions?''
  \emph{Transportation Research Part D: Transport and Environment}, vol.~34,
  pp. 267--280, 2015.

\bibitem{Brockfeld.2004}
E.~Brockfeld, R.~D. K{\"u}hne, and P.~Wagner, ``Calibration and validation of
  microscopic traffic flow models,'' \emph{Transportation Research Record:
  Journal of the Transportation Research Board}, vol. 1876, no.~1, pp. 62--70,
  2004.

\bibitem{Souza.2021}
F.~de~Souza and R.~Stern, ``Calibrating microscopic car-following models for
  adaptive cruise control vehicles: Multiobjective approach,'' \emph{Journal of
  Transportation Engineering, Part A: Systems}, vol. 147, no.~1, 2021.

\bibitem{Li.2016}
L.~Li, X.~Chen, and L.~Zhang, ``A global optimization algorithm for trajectory
  data based car-following model calibration,'' \emph{Transportation Research
  Part C: Emerging Technologies}, vol.~68, pp. 311--332, 2016.

\bibitem{Ciuffo.2014}
B.~Ciuffo, V.~Punzo, and M.~Montanino, ``Global sensitivity analysis techniques
  to simplify the calibration of traffic simulation models. methodology and
  application to the idm car--following model,'' \emph{IET Intelligent
  Transport Systems}, vol.~8, no.~5, pp. 479--489, 2014.

\bibitem{Sobol.2001}
I.~M. Sobol', ``Global sensitivity indices for nonlinear mathematical models
  and their monte carlo estimates,'' \emph{Mathematics and Computers in
  Simulation}, vol.~55, no. 1-3, pp. 271--280, 2001.

\bibitem{Saltelli.2010}
A.~Saltelli, P.~Annoni, I.~Azzini, F.~Campolongo, M.~Ratto, and S.~Tarantola,
  ``Variance based sensitivity analysis of model output. design and estimator
  for the total sensitivity index,'' \emph{Computer Physics Communications},
  vol. 181, no.~2, pp. 259--270, 2010.

\bibitem{Punzo.2021}
V.~Punzo, Z.~Zheng, and M.~Montanino, ``About calibration of car-following
  dynamics of automated and human-driven vehicles: Methodology, guidelines and
  codes,'' \emph{Transportation Research Part C: Emerging Technologies}, vol.
  128, p. 103165, 2021.

\bibitem{Kerner.1996}
B.~S. Kerner and H.~Rehborn, ``Experimental features and characteristics of
  traffic jams,'' \emph{Physical review. E, Statistical physics, plasmas,
  fluids, and related interdisciplinary topics}, vol.~53, no.~2, pp.
  R1297--R1300, 1996.

\end{thebibliography}
%
%

\addtolength{\textheight}{-12cm}   

\end{document}